\documentclass{article}

\usepackage[preprint,nonatbib]{neurips_2020}

\usepackage[utf8]{inputenc} 
\usepackage[T1]{fontenc}    
\usepackage{algorithm}
\usepackage{algorithmic}
\usepackage{amsmath, amssymb}
\usepackage{amsfonts}       
\usepackage{bm}
\usepackage{booktabs}
\usepackage{dsfont}
\usepackage{graphicx}
\usepackage{graphbox}
\usepackage{hyperref}
\usepackage{mathtools}
\usepackage{microtype}      
\usepackage{multirow}
\usepackage{nicefrac}       
\usepackage[separate-uncertainty=true,multi-part-units=repeat]{siunitx}
\usepackage{pifont}
\newcommand{\cmark}{\ding{51}\xspace}
\newcommand{\xmark}{\ding{55}\xspace}
\usepackage{url}
\usepackage{xcolor}
\usepackage{xspace}

\usepackage[numbers,compress]{natbib}

\hypersetup{
  pdfinfo={
    Title={Latent Domain Learning with Dynamic Residual Adapters},
    Author={Lucas Deecke, Timothy Hospedales, Hakan Bilen}
  }
}

\definecolor{blue}{HTML}{2778B1}
\newcommand{\bluebullet}{\textcolor{blue}{\textbullet}}
\definecolor{green}{HTML}{3AA039}
\newcommand{\greenbullet}{\textcolor{green}{\textbullet}}
\definecolor{orange}{HTML}{FC8126}
\newcommand{\orangebullet}{\textcolor{orange}{\textbullet}}
\definecolor{purple}{HTML}{6A3997}
\newcommand{\purplebullet}{\textcolor{purple}{\textbullet}}
\definecolor{red}{HTML}{E02025}
\newcommand{\redbullet}{\textcolor{red}{\textbullet}}
\definecolor{venom}{HTML}{00FC33}

\makeatletter
\DeclareRobustCommand\onedot{\futurelet\@let@token\@onedot}
\def\@onedot{\ifx\@let@token.\else.\null\fi\xspace}
\def\cf{c.f\onedot} 
\def\eg{e.g\onedot} 
\def\ie{i.e\onedot} 
 
\def\etc{etc\onedot}

\def\ours{DRA\xspace} 
\makeatother

\newcommand{\B}{\bfseries} 

\title{Latent Domain Learning with Dynamic \\ Residual Adapters}

\author{%
  Lucas Deecke, Timothy Hospedales, Hakan Bilen \\
  School of Informatics \\
  University of Edinburgh \\
  \texttt{\{l.deecke,t.hospedales,hbilen\}@ed.ac.uk} \\
}

\begin{document}

\maketitle

\begin{abstract}
A practical shortcoming of deep neural networks is their specialization to a single task and domain. While recent techniques in domain adaptation and multi-domain learning enable the learning of more domain-agnostic features, their success relies on the presence of domain labels, typically requiring manual annotation and careful curation of datasets. Here we focus on a less explored, but more realistic case:~learning from data from multiple domains, without access to domain annotations. In this scenario, standard model training leads to the overfitting of large domains, while disregarding smaller ones. 
We address this limitation via \textit{dynamic residual adapters}, an adaptive gating mechanism that helps account for latent domains, coupled with an augmentation strategy inspired by recent style transfer techniques. Our proposed approach is examined on image classification tasks containing multiple latent domains, and we showcase its ability to obtain robust performance across these. Dynamic residual adapters significantly outperform off-the-shelf networks with much larger capacity, and can be incorporated seamlessly with existing architectures in an end-to-end manner.
\end{abstract}

\section{Introduction}
\label{sec:introduction}

While the performance of deep learning has surpassed that of humans in a range of tasks, machine learning models perform best when learning objectives are narrowly defined. Practical realities however often require the learning of joint models over semantically different information. In this case, best performances are usually obtained by fitting a collection of models, with each model solving an individual sub-task. This is somewhat disappointing, seeing how humans and other biological systems are capable of flexibly adapting to a large number of scenarios.

Past solutions that address this problem tend to fall into some category of multi-domain learning, where -- different from the broader multi-task scenario -- a single loss function is shared across tasks. Multi-domain learning however relies firmly on the availability of domain annotations, for example used to control modules in domain-specific architectures \citep{rebuffi18,liu19}.

The reliance on domain annotations is however not limited to the multi-domain scenario, their presence is required in adversarial domain adaptation \citep{ganin16}, or multi-source domain adaptation \citep{xu18,peng19a}. Other examples of dependence on domain labels can be found in the continual learning literature, where task-specific memories guide the learning process \citep{lopez-paz17,18}, in domain generalization \citep{li18,li19a,li19b}, or meta learning \citep{finn17}.

\begin{table*}[h]
\centering
\caption{Domain-wise and weighted accuracies for 9x ResNet26 learned individually on each domain, versus a single ResNet26 that learns on all domains jointly. \cmark and \xmark denote presence or absence of domain annotations $d$. $\pi_d$ indicates the overall share of each domain; on smaller ones (\eg Aircraft) performance losses are significant.} \label{table:introduction}
\vspace*{.1cm}
\resizebox{\columnwidth}{!}{
\begin{small}
\begin{tabular}{ccccccccccccc}
\toprule
                        & $d$               & Airc.    &  C-100  & Daim. & Dtd    & Gtsrb & Omn.  & Svhn  & Ucf101 & Vgg-F.  & Avg.    \\
\midrule
$\pi_d$                 &                   & 0.052    &  0.156  & 0.091 & 0.028  & 0.122 & 0.1   & 0.406 & 0.016  & 0.03    &         \\
9x ResNet26             & \cmark            & 39.48    &  77.96  & 99.95 & 38.19  & 99.95 & 87.62 & 95.12 & 73.00  & 65.20   & 87.01   \\
ResNet26                & \xmark            & 31.35    &  70.71  & 99.49 & 33.67  & 99.87 & 87.80 & 94.64 & 58.25  & 60.39   & 84.73   \\
\midrule
Relative [\%]           &                   & -20.59   &  -10.25 & -0.46 & -13.42 & -0.08 & 0.21  & -0.51 & -25.32 & -7.96   & -2.69   \\
\bottomrule
\end{tabular}
\end{small}
}
\end{table*}

The above approaches are successful when domain annotations are available. In the real world however, these can be difficult or expensive to obtain. Consider images that were scraped from the web. Existing multi-domain approaches would require that these scraped images are further annotated for the mixture of content types they will necessarily contain, such as real world images or studio photos \citep{saenko10}, clipart or sketches \citep{li17}.


In this paper we relax this requirement, and consider the alternative scenario of latent domain learning. This encompasses any task where we have reason to believe that some partitioning of the data  would make sense, but we are uncertain about what a good partitioning might look like, or have inadequate resources to label all data. And as our experiments show, even when domain labels exist, there is no guarantee that these were chosen optimally, nor that learning them isn't the better option.

Learning on latent domains poses a major problem for standard deep learning approaches, which have a strong tendency to overfit to large modes in data. This issue is displayed in Table \ref{table:introduction}: the first row shows performance of 9x ResNet26 architectures \citep{he16} individually finetuned to all datasets in Visual Decathlon \citep{rebuffi17}.\footnote{Except ImageNet \citep{deng09}, which we omit due to its overweight.} The second row shows the performance of a single ResNet26 jointly learned on all these tasks. While the overall loss in weighted accuracy appears modest (-2.69\%), domain-conditional accuracies reveal that the single model achieves this feat by overaccounting for large datasets (Svhn, Omniglot, \etc), while disregarding smaller ones (Aircraft, Dtd, Ucf101).


We illustrate the difference between multi-domain and latent domain learning scenarios via graphical models in Figure \ref{fig:graphical-model_dra}, and further discuss this difference in Section \ref{ssec:problem_setting}. We subsequently propose novel mechanisms designed to address the central issues in latent domain learning: dynamic residual adapters (Section \ref{ssec:dyn_ra}) allow us to achieve robust performance metrics on small domains, all without trading performance on larger ones (see experiments in Section \ref{sec:experiments}). Our proposed module is efficient, can be incorporated and trained seamlessly with existing architectures, and is able to surpass the performance of domain-supervised approaches that rely on human-annotated data (Section \ref{ssec:pacs}).

\section{Related work}
\label{sec:related_work}

Multi-domain learning relates most closely to our paper. The state-of-the-art introduces small convolutional corrections in residual networks to account for individual domains \citep{rebuffi17,rebuffi18}. \citet{stickland19} extend this approach to obtain efficient multi-task models for related language tasks. Other recent work makes use of task-specific attention mechanisms \citep{liu19}, attempts to scale task-specific losses \citep{kendall18}, or addresses tasks individually at the level of gradients \citep{chen17}.

A lack of domain annotations has recently attracted interest in unsupervised domain adaptation. \citet{mancini18} estimate batch statistics of domain adaptation layers with Gaussian mixture models using only few domain labels. \citet{peng19b} study the shift from some source domain to a target distribution that contains multiple latent domains. In our setting, there is no shift between source and target distributions, instead the focus lies on learning parameter efficient models that generalize well across multiple latent domains simultaneously.

Previous work that directly addresses latent domain learning can be found in \citet{xu14} which use exemplar SVMs to account for latent domains and thereby generalize to new ones, while \citet{xiong14} study the discovery of latent domains by clustering via maximization of mutual information.

Our work connects to  multi-modal learning: \citet{chang18} propose an architecture for person reidentification that accounts for modality in distributions. \citet{deecke18} normalize data in separate batches to account for differences in feature distributions. Furthermore, our work is loosely related to learning universal representations \citep{bilen17}, which \citet{tamaazousti19} use as a guiding principle in designing more transferable models.

Dynamic architectures have recently attracted considerable attention, with solutions from reinforcement learning \citep{zoph16,pham18} or using Bayesian optimization \citep{kandasamy18}. For differentiable dynamic architectures, two components are commonly used: Gumbel-softmax sampling \citep{jang16}, \eg leveraged in dynamic computer vision architectures \citep{veit18,sun19}, and mixtures of experts \citep{jacobs91,jordan94}, used to scale deep learning models to large problems spaces \citep{shazeer17}, and for universal object detection \citep{wang19}.

From the perspective of algorithmic fairness, a desirable property for models is to ensure consistent predictive equality across different identifiable subgroups in the data \citep{zemel13,hardt16,fish16,corbett-davies17}. This relates to the central goal in latent domain learning, which is to ensure robustness on small latent domains.

\section{Method}
\label{sec:method}


\subsection{Problem setting}
\label{ssec:problem_setting}

In multi-domain learning, it is assumed that data is sampled i.i.d. from some mixture of domain distributions $\mathbb P_d$, with domain labels $d=1,\dots,D$. Together, they constitute the underlying distribution as $\mathbb P = \sum_d \pi_d \mathbb P_d$, where each domain is associated with a relative share $\pi_d = n_d/n$, with $n$ the total number of samples, and $n_d$ those belonging to the $d$'th domain. The target space is usually made up of mutually exclusive classes of the underlying domains, \ie $| \mathcal Y | = \sum_d | \mathcal Y_d |$. In multi-domain learning, the domain label $d$ is always available.

While the two are closely related, in the broader multi-task scenario the nature of underlying tasks $t=1,\dots,\mathcal T$ is inherently different, and learning on each task distribution $\mathbb P_t$ is associated with an individual loss function $L_t$ (for example, one task may be object classification, the other semantic segmentation). When the $L_t$ are equivalent across tasks, multi-task learning and multi-domain learning coincide.

Different from learning in a traditional multi-domain setting, in latent domain learning the information associating each sample to a domain $d$ is no longer available. As such, we cannot infer labels $y$ from sample-domain pairs $(x,d)$ and are instead forced to learn a single model $f_\theta$ only from $x$. 


A common baseline in multi-domain learning is to finetune a set of $D$ individual models, one for each domain. The task in the multi-domain literature is then to overcome this baseline \citep{rebuffi18}. We use this particular baseline throughout the paper, and show that in some cases, even when domain annotations were chosen carefully, a domain-unsupervised approach can surpass the performance of domain-supervised ones, see Section \ref{ssec:pacs_results}.


\begin{figure}[t]
\begin{center}

\centerline{
\includegraphics[width=\columnwidth]{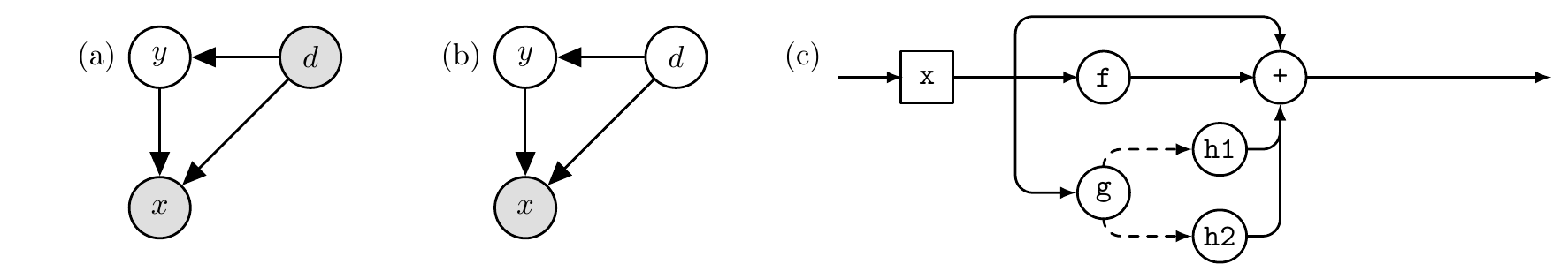}
}
\end{center}
\caption{Graphical models for (a) multi-domain learning, (b) latent domain learning. (c) A ResNet block, equipped with a dynamic residual adapter ($K=2$). Incoming samples \texttt{x} pass down three streams: an identity function, a transformation via a large convolution \texttt{f}, as well as an evaluation by expert gates \texttt{g}, which dynamically assigns (dashed arrows) small corrections \texttt{h1} and \texttt{h2}.} \label{fig:graphical-model_dra}
\end{figure}

\subsection{Residual adaptation}
\label{ssec:ra}

A widely adopted approach in multi-domain learning builds on the assumption that features from large pretraining tasks are universal, and only small convolutional transformations at each layer of the network are needed to correct for domain-specific differences. \citet{rebuffi17,rebuffi18} use this insight to extend the layer-wise transformation of the widely adopted residual architecture \citep{he16}:

\begin{equation}
x + f_{\theta}(x) + h_{\alpha,d}(x),
\end{equation}

where $f_{\theta}$ are the main convolutions of the residual network, and $h_{\alpha,d}$ are light-weight (\ie $|\alpha| \ll |\theta|$) convolutional corrections. In this work access to $d$ is removed, resulting in two new challenges: we have no a priori information about the right number of corrections $\{h_{\alpha,d}\}$, and we cannot use $d$ to decide which one of these to apply. Throughout the next section, we present an alternative strategy of inspecting $x$ and choosing relevant corrections $h_{\alpha}$ on the fly.

\subsection{Dynamic residual adapters}
\label{ssec:dyn_ra}

While there is no access to domain labels $d$ in latent domain learning, we may still assume $\mathbb P$ is constituted by several domain distributions $\mathbb P_d$. In order to account for these in a fully domain-unsupervised fashion, we propose the use of dynamic residual adapters (\ours): each incoming sample $x$ is first processed by a set of corrections $\{h_{\alpha_{k}}\}_{k=1,\dots,K}$, which we parametrize with light-weight 1x1 convolutions. Next, a noisy mixtures of experts (MoE) gating mechanism $g_k$ is responsible for weighing each correction $h_{\alpha_k}$, under which $x$ is then transformed. In the $l$'th layer of the network, the subsequent feature representation computes as

\begin{equation}
x + f_{\theta_l}(x) + \sum_{k=1}^K g_{lk}(x) h_{\alpha_{lk}}(x),
\end{equation}

with $g_{lk}(x)$ the $k$'th component of the $l$'th gating function. For an illustration, see Figure \ref{fig:graphical-model_dra} (c). While we motivate \ours from learning on latent domains, note that many additional factors (\eg shape, pose, color) may enter the gate assignments as well. Extending networks with dynamic residual adapters exhibits strong performance on smaller modes, while retaining performance on larger ones, see our experiments in Section \ref{sec:experiments}.



We follow \citet{shazeer17} and parametrize the gating units via self-attention \citep{lin17}. The only learnable parameters are those of a small linear transformation $W\colon \mathcal C \to \mathbb R^K$, resulting in a light-weight parametrization of the gates as

\begin{equation}
g(x) = \text{Softmax}\{W^\intercal \phi(x) + \varepsilon \},
\end{equation}

where $\phi(x)$ is a projection onto the channel dimension $\mathcal C$ that averages out height and width (average pooling), $\varepsilon \sim \mathcal N(0,\Sigma_\varepsilon)$ a channel-wise exploration noise.\footnote{Exploration noise is fixed at $\Sigma_{ii}=\num{e-2} \, \forall i$, zero otherwise.} A final softmax ensures the gating mechanism corresponds to a valid categorical distribution over $K$ latent domains, i.e. $0 \leq g_k  \leq 1$ and $\sum_k g_k = 1$. How to choose $K$ is discussed in more detail in Section \ref{ssec:optimization}.

In practice, many other parametrizations of the gating function $g_k$ are possible. Self-attention gives rise to smooth assignments, allowing the weighted combination of different $h_{\alpha_{k}}$. Discrete assignments can be enforced through Gumbel-Softmax sampling \citep{jang16}. In practice, we found the latter approach to be too restrictive, and a smooth interpolation via MoE to be the more favorable option. We compare gate parametrizations in Table \ref{table:ablation} of the Appendix.

\begin{table*}[t]
\centering
\caption{Performance of 9x ResNet26 individually finetuned to all domains, versus ResNet26, ResNet56 and our dynamic residual adapters with $(K,\eta)=(2,0.025)$. Best latent domain performance highlighted.} \label{table:experiments_1}
\vspace*{.1cm}
\resizebox{\columnwidth}{!}{
\begin{small}
\begin{tabular}{cccccccccccc}
\toprule
                        & $d$    & Airc.    &  C-100  & Daim. & Dtd    & Gtsrb & Omn.  & Svhn  & Ucf101 & Vgg-F.  & Avg.    \\
\midrule
$\pi_d$                 &        & 0.052    &  0.156  & 0.091 & 0.028  & 0.122 & 0.1   & 0.406 & 0.016  & 0.03    &         \\
9x ResNet26             & \cmark & 39.48    &  77.96  & 99.95 & 38.19  & 99.95 & 87.62 & 95.12 & 73.00  & 65.20   & 87.01   \\
\midrule
ResNet26                & \xmark & 31.35    &  70.71  & 99.49 & 33.67  & 99.87 &\B87.80& 94.64 & 58.25  & 60.39   & 84.73   \\
ResNet56                & \xmark & 34.62    &  71.63  &\B99.52& 34.79  &\B99.90& 87.72 &\B95.12& 60.66  & 57.55   & 85.22   \\
Ours                    & \xmark &\B38.28   &\B78.16  & 99.13 &\B40.64 & 99.77 & 86.61 & 94.17 &\B63.88 &\B69.80  &\B86.46  \\
\bottomrule
\end{tabular}
\end{small}
}
\end{table*}

\subsection{Style exchange augmentation}
\label{ssec:style_exchange}

The central challenge in latent domain learning is the tendency of large domains in $\mathbb P$ to suppress smaller ones. Besides accounting for this via dynamic residual adapters, we introduce an augmentation technique that encourages information exchange between domains.

We are motivated by the following example: assume two classes (say, cats and dogs) each with two latent domains (sketches and photos). Ideally, we would want to encourage the model to learn a domain-agnostic representation of $x$, from which it may infer $y$, invariant of $d$. We achieve this here by augmenting $x$ with the style information of a second sample $z$, drawn at random from $\mathbb P$ (so potentially, but not necessarily crossing domains).


To ensure that this can be done with small computational overhead, we augment samples after they have been compressed into a dense representation, \ie after the last convolutional layer of the network. Formally, we factorize the model (with convolutional parameters $\theta$ and $\alpha$) and subsequent classifier into $f_{\theta\alpha} = V^\intercal t_{\theta\alpha}$. As shorthand for a sample's final representation we denote $a_x = t_{\theta\alpha}(x)$, and map the low-level feature representation of the pertubation $z$ onto the target $x$ via

\begin{equation}
\eta\Big[\sigma_{a_z} \Big( \frac{a_x - \mu_{a_x}}{\sigma_{a_x}} \Big) + \mu_{a_z} \Big] + (1-\eta)a_x,
\end{equation}

borrowing from recent work in the style transfer literature \citep{huang17}. $\eta \in [0,1]$ is introduced to scale the augmentation (higher values will augment more aggressively), moments $\mu, \sigma$ are estimated across channel, height and width of $a_x$ and $a_z$, respectively.

In our experiments, we randomly pair samples in each mini batch. We find that modest values for $\eta$ work best, as augmenting to aggressively risks breaking the relationship between $x$ and its corresponding label $y$, see Figure \ref{fig:style_transfer} in the Appendix. When learning on latent domains, exchanging feature-level style information between samples works much better in practice than applying other recently proposed generic approaches, \eg MixUp \citep{zhang17}, \cf the ablation in Table \ref{table:ablation}.

\section{Experiments}
\label{sec:experiments}

We consider two experimental settings to evaluate our proposed approaches: a traditional multi-domain scenario but without access to domain information (Section \ref{ssec:multi_domain}), and a single dataset that contains multiple latent domains (Section \ref{ssec:pacs}).


\subsection{Latent multi-domain}
\label{ssec:multi_domain}

The first trial combines nine datasets from the Visual Decathlon challenge \citep{rebuffi17} that contain a variety of different images, with mutually exclusive labels, \ie $|\mathcal Y| = \sum_d |\mathcal Y_d|$.\footnote{For SVHN and CIFAR-100, this would for example give rise to a 110-dimensional label space. Unlike in multi-domain learning however, in latent domain learning models may erroneously classify samples from CIFAR-100 to SVHN, and vice versa.} Note the goal here is not to compare to the performance of domain-supervised approaches that Visual Decathlon was designed for, but to show that deep networks struggle with incorporating small latent domains when no domain annotations are provided.

\subsubsection{Optimization}
\label{ssec:optimization}

Initial ResNet parameters were obtained from pretraining on ImageNet \citep{deng09}. For dynamic residual adapters, only gates and corrections are learned, the ResNet26 backbone remains fixed at its initial parameters. This requires less parameters to be learned (see Figure \ref{fig:memory_activations}, left), while also benefiting performance (\cf Table \ref{table:ablation}). We followed the exact same optimization routine across models and experiments: we trained for 120 epochs using stochastic gradient descent (momentum parameter of 0.9), batch size of 128, weight decay of \num{e-4}, and initial learning rate of 0.1 (reduced by 1/10 at epochs 80, 100). For dataset splits, we followed \citet{rebuffi17}. Test accuracies displayed in tables were obtained by averaging over results from five random initializations.

We experimented with different values for the amount of style exchange. While a range of values improve over having no augmentation, the best results were obtained by limiting this to a modest amount (results shown use $\eta=0.025$ throughout), see Figure \ref{fig:style_transfer}. Setting $K=2$ provides the network with $2L$ corrections (where $L$ denotes the number of layers), which was sufficient to achieve robust performance across latent domains. There is a small but limited performance gain from increasing $K$ further, see Section \ref{ssec:pacs}.

\begin{table*}[t]
\centering
\caption{Results on the PACS dataset. Shown are performances for ResNet26, MLFN, a domain-supervised ensemble of 4x ResNet26, and \ours (with $K=2, 4$). Third column lists the number of parameters that have to be learned in each approach.} \label{table:pacs}
\vspace*{.1cm}
\begin{small}
\begin{tabular}{ccccccccc}
\toprule
                       & $d$    & Param.[$\approx$]& Art Painting & Cartoon & Photo   & Sketch & Avg.    \\
\midrule
$\pi_d$                &        &                  & 0.205        & 0.235   & 0.167   & 0.393  &         \\
RA \citep{rebuffi18}   & \cmark & 2.6 mil          & 86.47        & 92.37   & 95.15   & 94.61  & 92.51   \\
4x ResNet26            & \cmark & 24.8 mil         & 88.77        & 95.97   & 95.95   &\B95.83 & 94.44   \\
ResNet26               & \xmark & 6.2 mil          & 84.42        & 94.66   & 94.98   & 95.42  & 92.91   \\
MLFN \citep{chang18}   & \xmark & 7.6 mil          & 78.50        & 91.29   & 89.97   & 93.20  & 89.20   \\
Ours, $K=2$            & \xmark & 1.4 mil          & 92.15        & \B96.62 & 97.09   & 95.34  & 95.28   \\
Ours, $K=4$            & \xmark & 2.8 mil          & \B92.75      & 95.97   & \B97.73 & 95.53  &\B95.43  \\
\bottomrule
\end{tabular}
\end{small}
\end{table*}

\subsubsection{Results}

For a domain-supervised baseline, we finetune 9x ResNet26 \citep{he16} (four layers in each block; channel widths of 64, 128, 256), one individual model for each latent domain $\mathbb P_d$. We then use the exact same model to learn a joint classifier on $\mathbb P$. Next, we couple dynamic residual adapters with the very same ResNet26. For further comparison, we also include a significantly deeper ResNet56. Results are shown in Table \ref{table:experiments_1}.

Loosing access to domain labels significantly harms performance (9x ResNet26 versus ResNet26). While performance on larger domains is not impacted, the performance on smaller domains (Dtd, Ucf101, Vgg-Flowers) suffers considerably.

This problem is not addressed by simply increasing the depth of the network: while weighted accuracy increases slightly, a ResNet56 suffers the same issues, leaking performance on small domains. Without ever having access to domain annotations, the flexible assignments of convolutional corrections in our dynamic residual adapters close a large portion of this gap. For further analysis on how this is achieved, see Section \ref{ssec:gate_activations} and \ref{ssec:gate_paths}.

\subsubsection{Memory requirements}
\label{ssec:memory}

\begin{figure}[t]
\begin{center}
\centerline{
\includegraphics[width=.47\columnwidth]{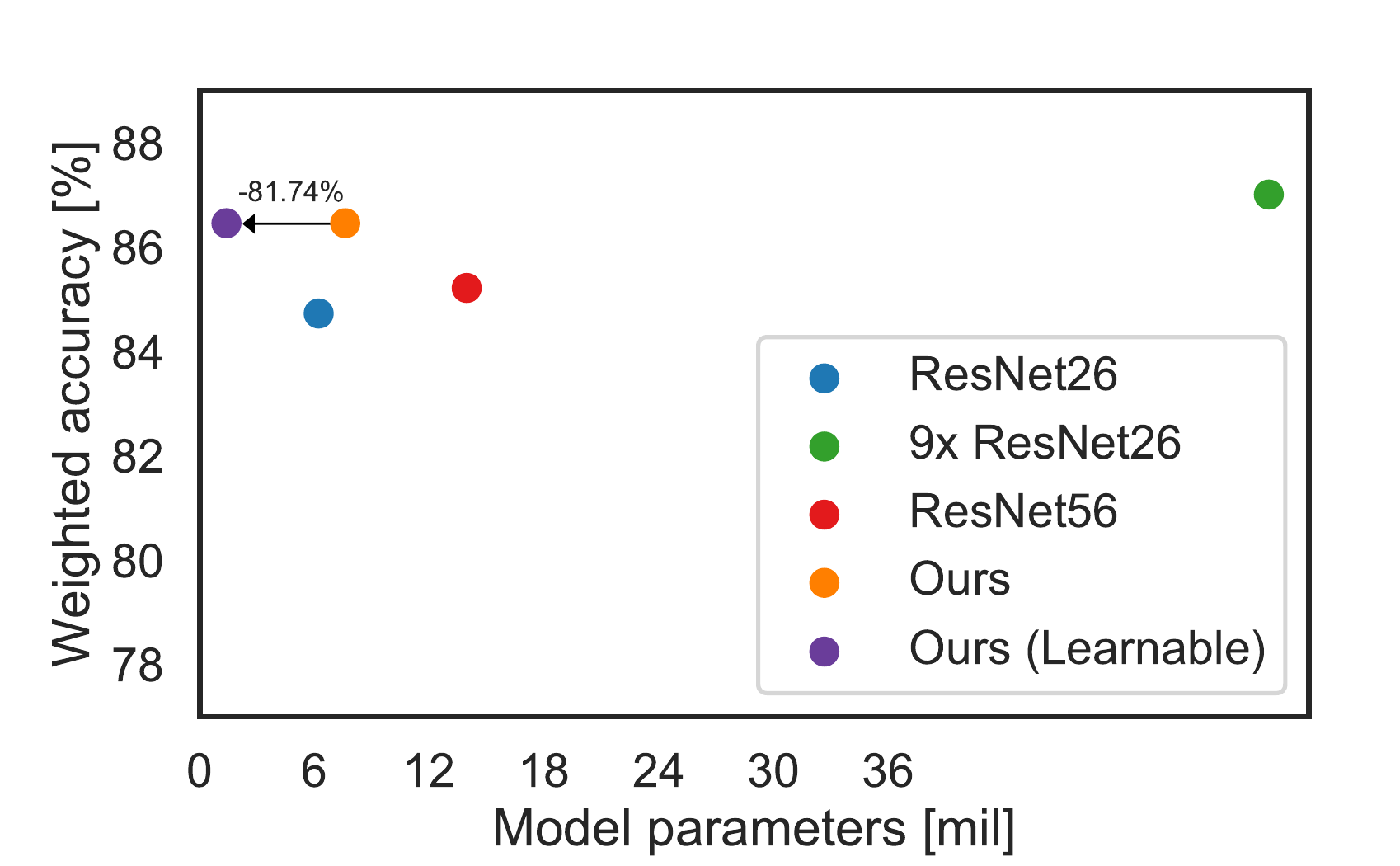}
\hspace*{.2cm}
\includegraphics[width=.47\columnwidth]{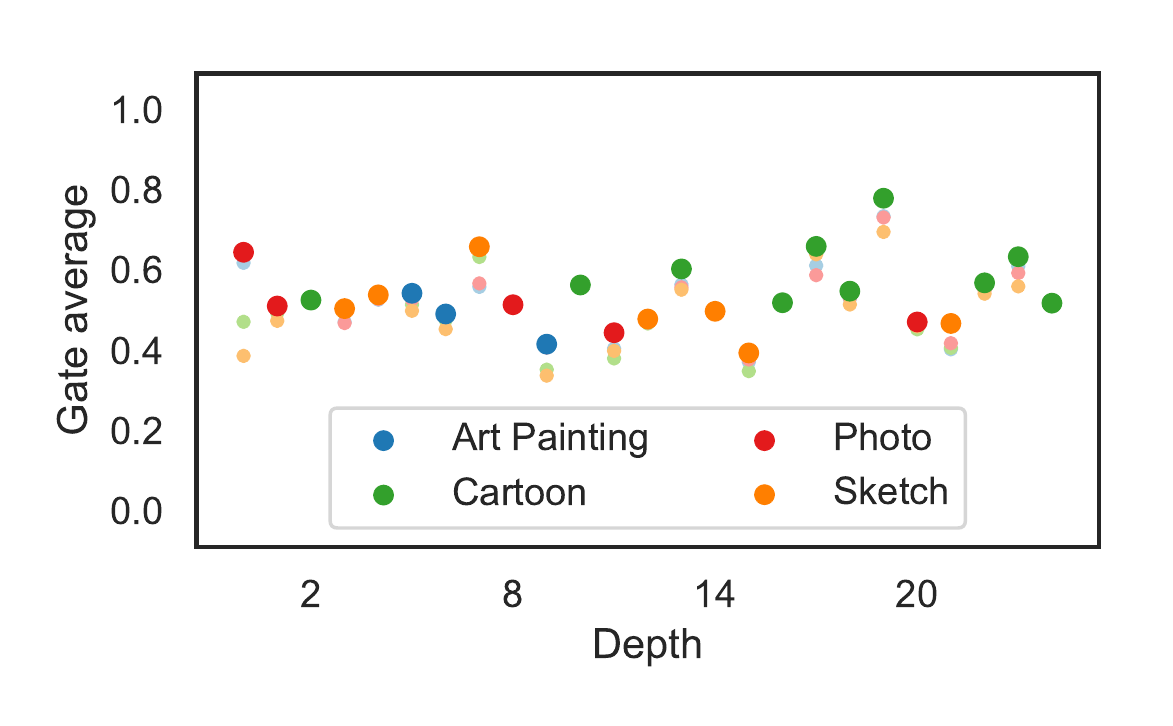}
}
\end{center}
\vspace*{-.6cm}
\caption{Left:~memory requirements of ResNet26 (\bluebullet), ResNet56 (\greenbullet), and 9x ResNet26 (\redbullet). For our dynamic residual adapters (\orangebullet), only a small portion of parameters need to be learned (\purplebullet). Right:~Activations of the gating mechanism $g$ for samples from each domain distribution $\mathbb P_d$ at different layers of the network. The domains that activate each gate strongest are highlighted.} \label{fig:memory_activations}
\end{figure}

Adding corrections $h_{\alpha_k}$ to the residual network results in additional memory requirements.\footnote{At each layer $\mathcal O (K|\mathcal C| + K|\mathcal C|^2)$ learnable parameters are needed to parametrize gates and corrections, respectively.} We show in Figure \ref{fig:memory_activations} (left) that these are very modest. In particular when comparing the number of learnable parameters, dynamic residual adapters have a significant advantage: the pretrained convolutions of the ResNet stay fixed throughout, only the gates and corrections have to be learned, amounting to less than 20\% of parameters in the network.

\subsection{Joint label space}
\label{ssec:pacs}


The second trial examines performance on a dataset called PACS \citep{li17}, standing for its four constituting domains (\textit{art painting, cartoon, photo, sketch}; examples shown in Figure \ref{fig:pca_qualitative}, right). Each domain contains samples of equivalent classes (``giraffe'', ``guitar'', \etc). The domains are unbalanced (see $\pi_d$ in Table \ref{table:pacs}).

We reserved 20\% of samples for evaluation, leaving the remainder for training. Splits were computed at random, as we assume no a priori knowledge of domain memberships. We make no changes to the optimization described in Section \ref{ssec:optimization}.

\subsubsection{Results}
\label{ssec:pacs_results}

The results in Table \ref{table:pacs} show that dynamic residual adapters improve considerably over a single ResNet26 baseline. While the largest domain \textit{sketch} is handled well by the traditional model, dynamic residual adapters can much better account for the small domains that also constitute $\mathbb P$.

To the best of our knowledge, latent domain learning has not been targeted through customized deep learning architectures. A related baseline is MLFN \citep{chang18}, which builds on ResNeXt \citep{xie17} to define a latent-factor architecture that accounts for multi-modality in data. Crucially, where we share small convolutional corrections at every layer, MLFN instead enables and disables entire network blocks. Arguably, our more fine-grained approach to parameter sharing allows us to outperform MLFN.

We also evaluate domain-supervised residual adapters \citep{rebuffi18}. While these have been shown to work extremely well in the multi-domain scenario, their performance here was sub-par. This is likely a result of its per-domain corrections $h_{\alpha_d}$, which exhibit no cross-domain sharing of parameters. Dynamic residual adapters share parameters natively across domains: this substantially benefits performance on domains like \textit{art painting} (92.15\% versus 84.42\%), which shares a lot of visual information with \textit{photo} (\cf Figure \ref{fig:pca_qualitative}, left).

Lastly, we finetune 4x individual ResNet26 to each PACS domain for a strong domain-supervised baseline. Unsurprisingly, this outperforms the single ResNet26 trained jointly on all domains. While requiring only a fraction of learnable parameters (\num{1.4}{mil} versus \num{24.8}{mil}), DRA however surpasses the performance of the fully domain-supervised ensemble by sharing model parameters across domains.

We further examine the benefit of introducing larger numbers of residual corrections $\{ h_{a_k} \}_{k=1,\dots,K}$. While the performance edges up slightly, the sequential nature of the network arguably already allows complicated partitionings of the residual corrections for $K=2$, making larger $K$ unnecessary.


Domain-supervised approaches can be incorporated into DRA by setting $K=D$, and fixing gate assignments to $g_{lk}(x)=\mathds 1_{x\sim \mathbb P_k}$ across the network. This also encompasses any potential clustering of domains, which introduces a different, but fixed set of global gate activations. Evident from the results shown here, such global assignments are not always optimal, even when domains have been assigned as carefully as in PACS. Dynamic residual adapters remove the need to decide a priori what constitutes good domain separations, and instead dynamically share or separate features at every layer. We analyze how this occurs in additional detail in the next two sections, by looking at activation paths across the network.

\subsubsection{Gate activations}
\label{ssec:gate_activations}

In Figure \ref{fig:memory_activations} (right) we show average per-layer activations of the gating mechanism. Different domains activate gates at different depths of the network, while visually similar domains (such as \textit{art painting} and \textit{photo}, \cf examples in Figure \ref{fig:pca_qualitative}, left) tend to activate together. At some layers there is little need for multiple corrections $h_{\alpha_k}$. In this case, dynamic residual adapters simply relax to a uniform gate:~the activation becomes $g_k = 1/2 \,\, \forall k$, joining the unit into a single residual correction $h_{\alpha}$.

\subsubsection{Gate pathways}
\label{ssec:gate_paths}


\begin{figure}[t]
\begin{center}
\centerline{
\includegraphics[width=.55\columnwidth]{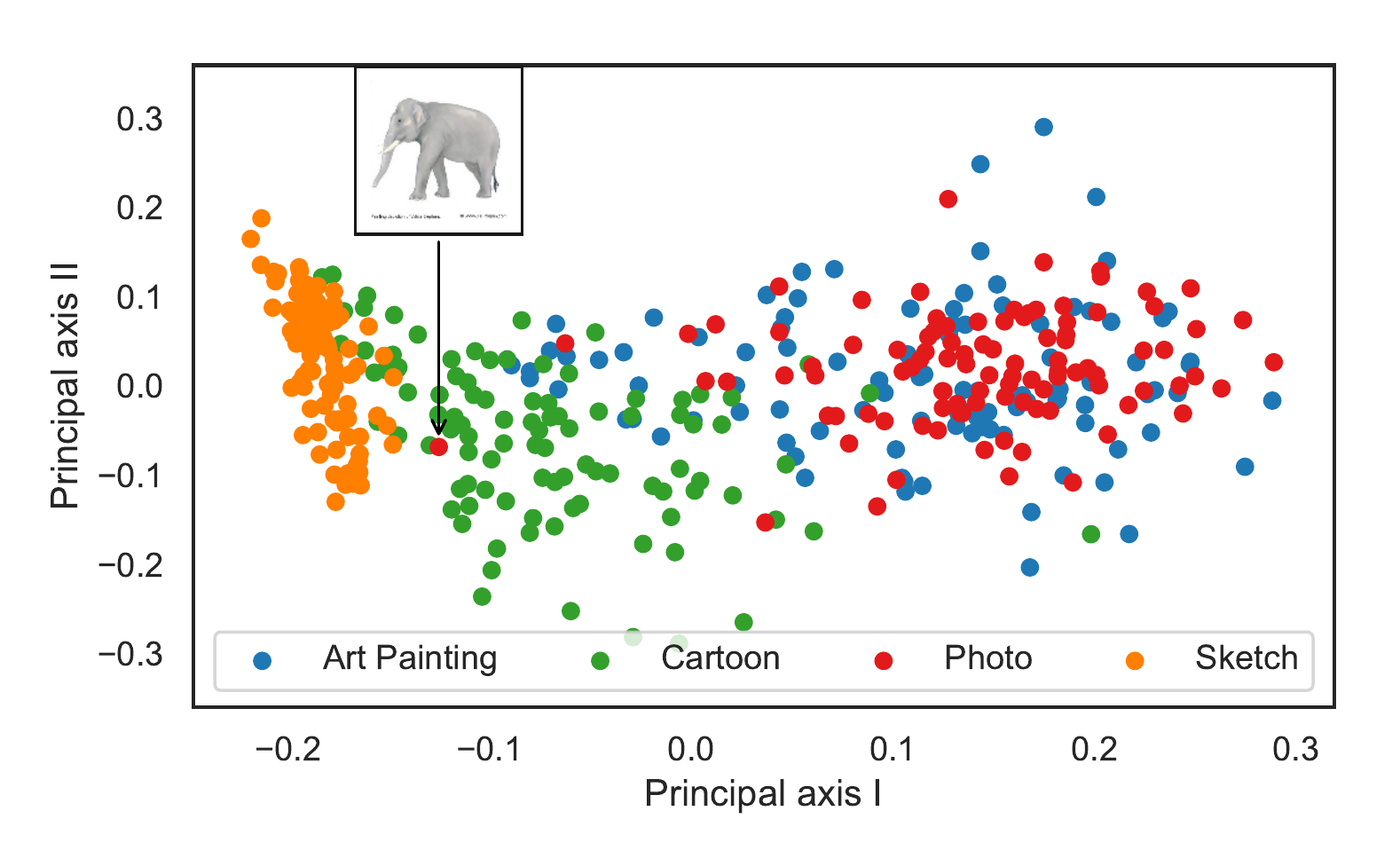} \hspace*{.2cm}
\raisebox{1.5cm}{\includegraphics[width=.15\columnwidth]{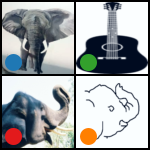}}
}
\end{center}
\vspace*{-.6cm}
\caption{Left:~PCA of samples represented by their $L$-dimensional activation paths. Gate paths are semantically meaningful: visually similar domains \textit{art painting} and \textit{photo} (\bluebullet,\redbullet) cluster together, \textit{cartoon} (\greenbullet) resides between real world imagery and \textit{sketches} (\orangebullet). A sample with an erroneous ground-truth domain label is highlighted. Right:~a group of samples (coloring corresponds to left-hand side) that share similar gate activation paths.} \label{fig:pca_qualitative}
\end{figure}

Intuitively, one might expect the gating mechanism to assign different convolutional corrections $h_{\alpha_k}$ to the different human-annotated domains in PACS. The visualization of gate activations in Figure \ref{fig:memory_activations} (right) seems to suggest the opposite:~the purity (share of maximum activation relative to all activations) is relatively low across the network.

Arguably, the above intuition oversimplifies how the network processes different samples, and is further contradicted by the performance loss that results from enforcing discreteness (\cf Gumbel-softmax in Table \ref{table:ablation}). To understand better how partitioning occurs, it is helpful to inspect what happens across the entirety of the network, and compare sets of feature activations throughout their processing. For a sample $x$, we define its $k$'th activation path across the $L$-layered network as

\begin{equation}
\xi_k(x) \triangleq \big(g_{k1}(x), \dots, g_{kL}(x)\big) \in \mathbb [0,1]^L.
\end{equation}

If samples have similar activations paths, this means they also share a large amount of parameters. As a group of samples with low pairwise distances in Figure \ref{fig:pca_qualitative} (right) shows, similar gate activations are indicative of visual similarity:~pose, color or edges of samples that group together are visibly related, compare in particular the pose of elephants from the \textit{photo} and \textit{sketch} domains.

We collected gate activation paths for (an equal number) of samples from all domains, and visualize their principal components in Figure \ref{fig:pca_qualitative} (left). This reveals an intuitive clustering of domains: visually similar domains \textit{art painting} and \textit{photo} (\bluebullet,\redbullet) share a region. The manifold describing \textit{sketches} (\orangebullet) is arguably more primitive than those of the other domains, and indeed only maps to a small region. \textit{Cartoon} (\greenbullet) lies somewhere between sketches and real world imagery. Under visual inspection (\cf examples in Figure \ref{fig:pca_qualitative}, right) this makes perfect sense: a cartoon is, more or less, just a colored sketch. We highlight one particular elephant that resides amongst the \textit{cartoon} domain, but has been annotated as \textit{photo} in the PACS dataset. The ground-truth annotation is incorrect, but different from domain-supervised approaches, dynamic residual adapters are not irritated by this.



\section{Conclusion}


Recent work in multi-domain learning has been chiefly focused on a setting where domain annotations are assumed to be routinely available. As this requires careful curation of datasets, in real world scenarios this assumption can often be of limited merit. Dynamic residual adapters help inject adaptivity into networks, preventing them from overfitting to the largest domains in distributions, a common failure mode of traditional models. Not only does our approach successfully close a large amount of the performance gap to domain-supervised solutions, but in some scenarios -- even when domains have been assigned very carefully -- exceeds their performance.





\bibliography{literature}
\bibliographystyle{plainnat}

\clearpage

\section*{Appendix}

\subsubsection*{Ablation}

The ablations in Table \ref{table:ablation} show that latent domain learning benefits from both the addition of multiple dynamic residual adapters (\ie $K>1$) as well as style exchange augmentation between the feature maps of samples ($\eta>0$).

While adding a single residual adapter ($K=1$) helps the model with smaller modes, it registers performance losses on medium and larger ones. In line with what \citet{rebuffi17} report, when not fixing parameters $\theta$ of the ResNet convolutions, this leads to problems with overfitting.

\begin{figure}[h]
\begin{center}
\centerline{\includegraphics[width=.6\columnwidth]{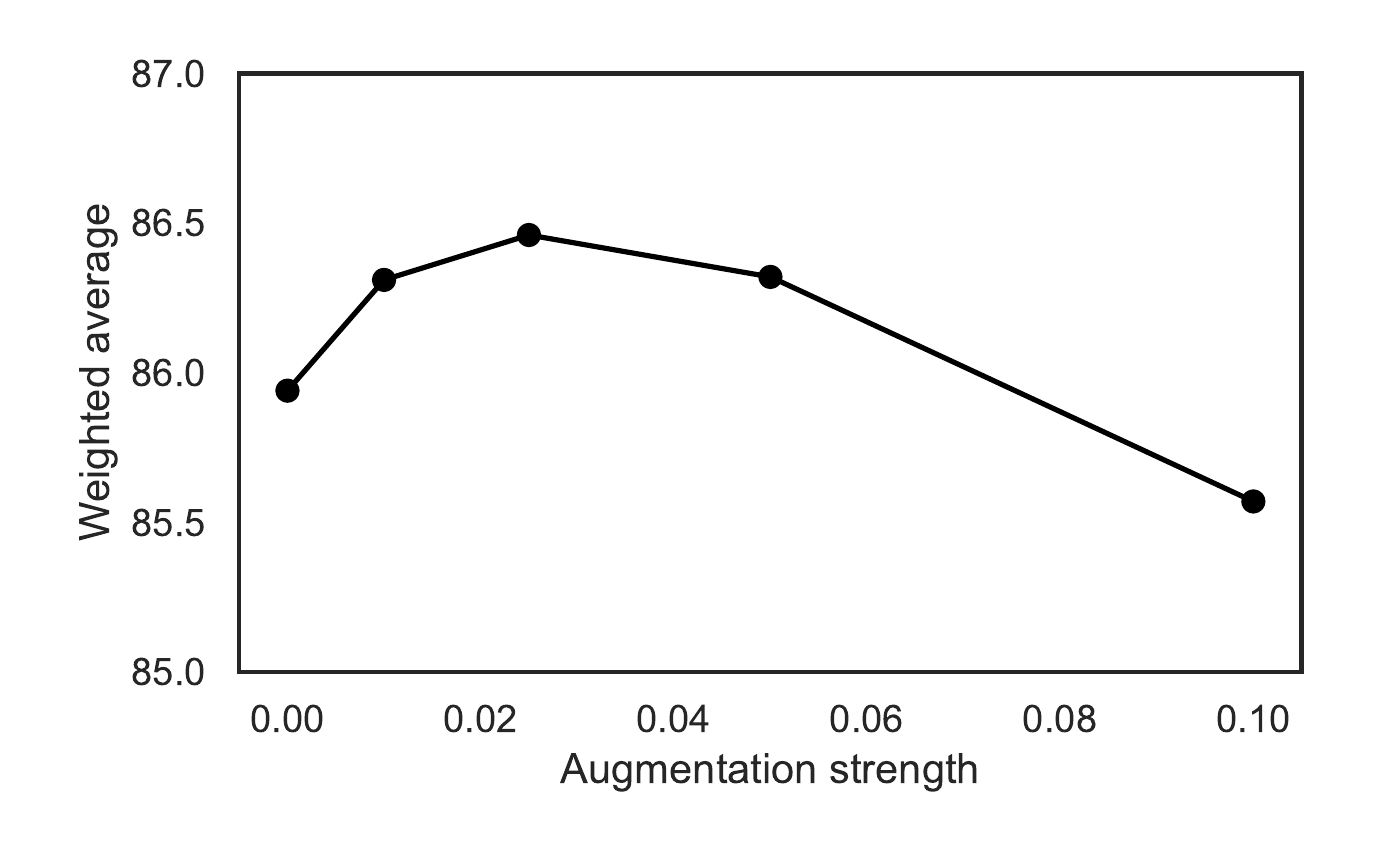}}
\end{center}
\vspace*{-.5cm}
\caption{Weighted average performance for DRA under different augmentation strengths $\eta$.} \label{fig:style_transfer}
\end{figure}

With no augmentation via style exchange ($\eta=0$), performance drops visibly. MixUp \citep{zhang17}, an alternative augmentation that interpolates between samples, is not equally well suited for latent domain learning. Results for additional choices of $\eta$ are shown in Figure \ref{fig:style_transfer}: modest values for $\eta$ work best, as augmenting too aggressively risks breaking the relationship between image-label pairs.

Replacing mixtures of experts with Gumbel-softmax sampling negatively impacted performance:~while some domains (CIFAR-100, Omniglot) are handled well, discrete gates struggle particularly with smaller ones. Performance for soft and straight-through Gumbel-softmax sampling was on par, and we report straight-through sampling here.


\begin{table*}[h]
\centering
\caption{An ablation study of our approach. First row uses a single ($K=1$) residual adapter in each layer, second row shows results when disabling style-exchange augmentation, \ie setting $\eta$ to zero. Third row finetunes all parameters, not just dynamic residual adapters. Fourth row couples dynamic residual adapters with MixUp. Fifth row contains results for a parametrization of the gates via straight-through Gumbel-softmax sampling. Results for dynamic residual adapters with $(K,\eta)=(2,0.025)$ shown last.} \label{table:ablation}
\vspace*{.1cm}
\resizebox{\columnwidth}{!}{
\begin{small}
\begin{tabular}{cccccccccccc}
\toprule
                        & Airc.    &  C-100  & Daim. & Dtd    & Gtsrb & Omn.  & Svhn  & Ucf101 & Vgg-F.  & Avg.    \\
\midrule
$K=1$                   & 36.24    &  75.33  & 98.89 & 37.87  & 99.67 & 86.68 & 94.02 & 60.04  & 68.43   & 85.65   \\
$\eta=0$                & 36.15    &  77.65  & 99.06 & 39.06  & 99.67 & 86.48 & 93.97 & 60.27  & 65.86   & 85.94   \\
Learn $\theta$          & 36.75    &  71.39  &\B99.47& 35.27  &\B99.85&\B87.32&\B94.73& 62.81  & 64.80   & 85.35   \\
MixUp                   & 30.66    &  67.40  & 97.09 & 36.49  & 99.72 & 86.03 & 93.41 & 56.25  & 65.78   & 83.47   \\
Gumbel                  & 33.21    &  76.43  & 98.88 & 37.45  & 99.64 & 86.83 & 93.67 & 60.81  & 69.31   & 85.56   \\
\midrule
Ours                    &\B38.28   &\B78.16  & 99.13 &\B40.64 & 99.77 & 86.61 & 94.17 &\B63.88 &\B69.80  &\B86.46  \\
\bottomrule
\end{tabular}
\end{small}
}
\end{table*}

\end{document}